\documentclass[twocolumn]{cinc}
\usepackage{mathtools, graphicx, xurl, hyperref, booktabs}
\usepackage{graphicx}

\title{Listen2YourHeart: A Self-Supervised Approach for Detecting Murmur in Heart-Beat Sounds}
\author{Aristotelis Ballas\textsuperscript{1},
Vasileios Papapanagiotou\textsuperscript{2},
Anastasios Delopoulos\textsuperscript{2},
Christos Diou\textsuperscript{1} \\ \ \\
\textsuperscript{1}Harokopio University of Athens, Athens, Greece \\
\textsuperscript{2}Aristotle University of Thessaloniki, Thessaloniki, Greece}

\begin{document}
\maketitle


\begin{abstract}
  Heart murmurs are abnormal sounds present in heartbeats, caused by turbulent
  blood flow through the heart. The PhysioNet 2022 challenge targets automatic
  detection of murmur from audio recordings of the heart and automatic detection
  of normal vs. abnormal clinical outcome. The recordings are captured from
  multiple locations around the heart.

  Our participation investigates the effectiveness of self-supervised learning
  for murmur detection. We train the layers of a backbone CNN
  in a self-supervised way with data from both this year's and the 2016
  challenge. We use two different augmentations on each training sample, and
  normalized temperature-scaled cross-entropy loss. We experiment with
  different augmentations to learn effective phonocardiogram
  representations. To build the final detectors we train two classification
  heads, one for each challenge task. We present evaluation results for all
  combinations of the available augmentations, and for our
  multiple-augmentation approach.

  Our team's, Listen2YourHeart, SSL murmur detection classifier received a weighted
  accuracy score of $0.737$ (ranked 13th out of 40 teams) and an outcome identification
  challenge cost score of $11946$ (ranked 7th out of 39 teams) on the hidden test set.
\end{abstract}


\section{Introduction}


Heart murmurs can be a sign of numerous serious heart conditions associated
with increased mortality
\cite{gbd_2017_causes_of_death_collaborators_global_2018}. The development of
accurate, non-invasive and easily accessible methods for heart murmur screening
can help with prevention and/or timely treatment, especially for populations with
reduced access to healthcare resources. The 2022 PhysioNet Challenge
\cite{Reyna2022.08.11.22278688, 9658215} attempts to incentivize researchers
towards developing methods for the automatic detection of abnormal heart waves
or heart murmurs, by analyzing phonocardiogram (PCG) signals. The analysis of
such auscultation signals can provide crucial information on the heart's
structure and potentially assist clinicians in early screenings or
assessments. We propose tackling the challenge through Self-Supervised Learning
(SSL) with Deep Neural Networks (DNN) for the classification of PCG signals.

Although DNNs have achieved remarkable success in various settings,
they have not yet been able to establish themselves in
clinical practice (including screening procedures).
There are several reasons that may contribute to this. First of all, a false
prediction may lead to serious consequences for the patient's health or may
incur unnecessary healthcare costs. Secondly, an already trained model may not
maintain its performance on newly collected data
\cite{9898255}. An additional constraint is
the fact that hospital and medical data may be weakly labeled, unlabeled and
may be affected by exogenous noise.

All the above add up to a need for generalizable, adaptable and
non-conventional model training methods. To this end, Self-Supervised Learning
(SSL) techniques have been proposed for biosignal classification and have
achieved promising results \cite{spathis_breaking_2022}.
In SSL, the main idea is to learn invariant representations by transforming or augmenting
the input data and by training a model to distinguish between representations originating
from the same signal against those that come from different ones.
In addition to medical image analysis \cite{CHEN2019101539}, SSL methods have also been
implemented for EEG \cite{gramfort2021learning} and ECG
\cite{sarkar2020self} signals. In
this work, we explore the usage of contrastive SSL \cite{jaiswal2020survey}
for the classification of PCG signals. Specifically, we (i) propose employing
SSL for the extraction of meaningful PCG signal representation, (ii) evaluate
our method's effectiveness with minimal fine-tuning on downstream tasks and
(iii) explore promising combinations of augmentations for the PCG signal
domain.

\section{Materials and methods}

The goal of the challenge \cite{Reyna2022.08.11.22278688, 9658215} is to
classify a patient on two classification tasks: murmur detection and clinical
outcome. For each patient there are one or more audio recordings available,
taken from 5 different locations (one of the four valves or other).

Our approach is split in two steps, and is similar for each classification
task. First, we train an audio-based classifier on windows of fixed length by
directly propagating the patient label to the windows (the only exception is for
the murmur class ``present'' where only windows from recording locations marked
as ``murmur locations'' are labeled as ``present''; windows from the rest
recording locations are labeled as ``absent''). Then, we predict a single label
for the patient by first aggregating the window-level labels to recording-level
labels and then the recording-level labels to a single patient-level label.

Our audio-based classifier is a convolutional neural network, based on the
architecture of \cite{papapanagiotou2017chewing}. However, given the limited
size of the dataset, we train the convolutional layers in a self-supervised way,
based on the approach of \cite{papapanagiotou2021self}. Thus, we also use the
dataset released for the 2016 challenge of PhysioNet \cite{Liu_2016} to extend
our training dataset. It should be noted that the 2016 dataset does not contain
labels that are relevant to the 2022 challenge.

The 2022 dataset contains audio at $4$ kHz while the 2016 dataset at $2$
kHz. Therefore, we choose to resample the 2022 dataset at $2$ kHz. We use windows of
$5$ sec length with an overlap of $2.5$ sec. However, the first and last $2$
seconds of each recording are discarded as they might contain transient noise
artifacts.

\subsection{Self-supervised training}

Each audio window $x$ is transformed into two versions, namely $x_1$ and $x_2$,
by applying two different sets of augmentations. Specifically, we examine the following
augmentations: cut-off filters (high-pass or low-pass, at $250$, $500$, and
$750$ Hz), rewinding the signal, inverting (multiply by $-1$), random scaling
(in the range of $0.5$ to $2.0$), adding zero-mean uniform noise and upsampling
by a factor of $2$, while cropping symmetrically to maintain window size.

To train the $5$ convolutional layers of the $5$-sec input window architecture
of \cite{papapanagiotou2017chewing} we project the output of the last
convolutional layer on an $128$-D space (via a fully connected layer) and then
minimize the normalized temperature-scaled cross-entropy loss
\cite{oord2018representation}. For the cosine similarity with
temperature \cite{hinton2015distilling}, we use a temperature value of $0.1$
based on some initial experiments on the training set and on
the findings of \cite{papapanagiotou2021self}.

Batch size is set to $256$ (leading to $512$ after the two augmentations) and
training is allowed for at most $50$ epochs (we use early stopping with a patience
of $5$ epochs on a validation subset, more on Section \ref{sec:exp_setup}). To
train such a large batch we employ the LARS optimizer \cite{you2017large} with a
linear warm-up of $5$ epochs to a learning rate of $0.1$ and then a cosine decay
(with an alpha of $1\%$).

Once training concludes, the projection head
is discarded, and the weights of the convolutional layers are frozen.

\subsection{Murmur classification}

Given a trained set of convolutional layers, we create a murmur classifier by
appending a classification head of $3$ FC layers
as in \cite{papapanagiotou2021self}. The last layer has
$3$ neurons, one per murmur class (``present'', ``unknown'', and ``absent'') and
uses softmax activation. We train only the FC layers using categorical
cross-entropy loss and the Adam optimizer, with a learning rate of $10^{-4}$.
Batch size is set to $64$ and we train for at most $200$ epochs using early stopping with
$20$ epochs patience on a validation subset (more on Section
\ref{sec:exp_setup}).

Training yields a murmur classifier for audio windows. Given an audio recording,
we follow the same procedure as in the training phase to extract windows
and apply the trained murmur classifier on each window
independently. We thus infer the probability of each murmur class for each
window. These probabilities are aggregated for the entire recording by averaging
across the windows (we have experimented with other aggregations but this method
evidently seems to be the most effective). Ultimately, we characterize the entire
recording with a single label, based on the most probable class.

To assign a ``present'' label to a patient, we require that at least one
recording has been labeled ``present''. If none of the recordings have been labeled as
such and at least one recording has been labeled as ``unknown'' the patient inherits
the ``unknown'' label. Finally, the ``absent'' label is assigned to the patient only if
all of the recordings have been labeled accordingly.

\subsection{Outcome classification}

Training and inference for clinical outcome are very similar to murmur training
and inference respectively. The only difference is that the classification
problem is now binary (``abnormal'' vs. ``normal'' clinical
outcome). Window-level aggregations are performed again using averaging, and we
require that at least one recording is classified as ``abnormal'' in order to
label the patient's clinical outcome as abnormal. If no such recording exists,
we label the patient as ``normal''.

\begin{figure*}
  \centering
  \includegraphics[width=0.85\textwidth]{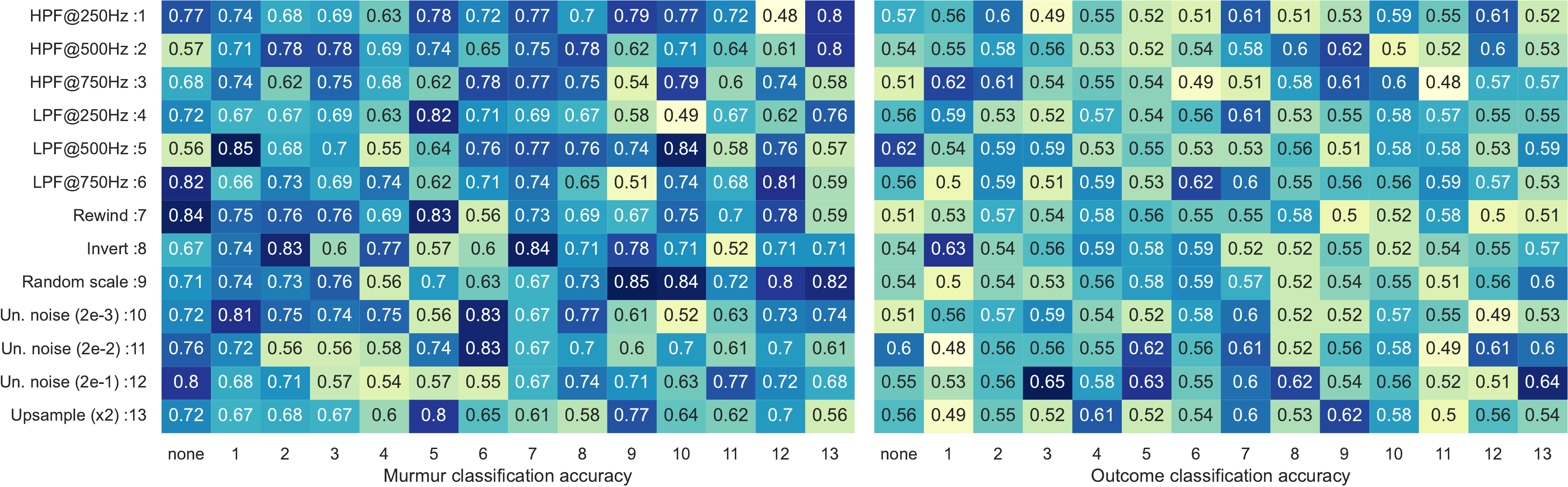}
  \caption{Augmentation comparison based on accuracy for murmur (left) and
    outcome (right) classification tasks.}
  \label{fig:augs}
\end{figure*}

\section{Experiments and results}

\subsection{Experimental setup}
\label{sec:exp_setup}

The dataset provided by the challenge contains data from $942$ patients. We
perform a stratified random split and keep $100$ patients as our test set,
denoted $S_\text{test}$ (different from the hidden test set of the
challenge). The remaining $842$ patients are again split (using a stratified
random split) into a training set $S_\text{train}$ and a validation set
$S_\text{val}$ ($80\%$-$20\%$). For SSL training, we also use the publically
available 2016 challenge dataset \cite{Liu_2016, physionet2016}, which contains
$3,541$ audio recordings, denoted $S_\text{16}$. We would like to note that
$S_{\text{16}}$ does not contain labels relevant to this year's
challenge. Thus, SSL training is done on $S_\text{16} \cup S_\text{train}$,
downstream training is done on $S_\text{train}$, and validation is always
$S_\text{val}$. All results are reported on $S_\text{test}$.

\subsection{Evaluation of augmentations}

To adequately explore the most promising augmentations, we first evaluate each
one of them separately (i.e. $x_1=f_i(x)$ and $x_2=x$ where $f_{i}$ are the
different augmentations we examine) and subsequently in pairs (i.e.  $x_1=f_i(x)$ and
$x_2=f_j(x)$). Figure \ref{fig:augs} summarizes the results for classification
accuracy both for murmur and outcome labels.

For murmur classification and a single augmentation, the first column of Figure
\ref{fig:augs} shows that a low-pass filter at $250$ Hz, rewind, and uniform
noise (of $0.02$ range) yield the best results (accuracy over $0.8$).
The highest accuracy for pairs is $0.85$ and is obtained by two combinations: a
high-pass filter at $500$ Hz with a high-pass filter at $250$ Hz, and random
scaling with itself. Both of these pairs have the same or similar
augmentations, however, and the resulting representations may not be
informative. The following combinations of diverse augmentations achieve $0.84$
accuracy: invert with rewind, low-pass filter at $500$ Hz with uniform noise
(range $0.002$), random scale with uniform noise (range $0.002$).

Accuracy for clinical outcome is generally lower. This is most probably due to
the fact that, in clinical practice, decision for clinical outcome does not
rely solely on listening to the audio recordings but also considers additional
information such as the patient/examinee's medical history (which were not
taken into account in our case). The best results are obtained by combining
uniform noise (range $0.01$) with either a high-pass filter at $750$ Hz,
yielding $0.65$ accuracy, or with upsampling by $2$, yielding $0.64$ accuracy.

\begin{table*}
  \centering
  \caption{Final results and baseline (fully supervised training on $S_{\text{train}}$).}
  \label{tab:results_final}
  \begin{tabular}{l|ccc|cccc}
    \toprule
    & \multicolumn{3}{c|}{\textbf{murmur}} & \multicolumn{3}{c}{\textbf{outcome}}\\
    & \textbf{F1-score} & \textbf{acc.} &  \textbf{w. acc.} & \textbf{F1-score} & \textbf{acc.} & \textbf{w. acc.} & \textbf{cost}\\\hline
    \midrule
    $f_3(f_2(f_1))$ vs. $g_2(g_1)$ & $0.544$ & $0.590$ & $0.668$ & $0.657$ & $0.590$ & $0.764$ & $10,541$ \\\hline
    $f_1$ vs. $g_2(g_1)$           & $0.606$ & $0.670$ & $0.700$ & $0.531$ & $0.570$ & $0.784$ & $10,892$ \\\hline
    $f_3(f_2(f_1))$ vs. $g_1$      & $0.561$ & $0.650$ & $0.700$ & $0.495$ & $0.540$ & $0.760$ & $11,513$ \\\hline
    \midrule
    Baseline                     & $0.524$ & $0.600$ & $0.558$ & $0.515$ & $0.550$ & $0.750$ & $11,276$ \\\hline
    \bottomrule
  \end{tabular}
\end{table*}

\subsection{Challenge results}

Based on these findings, we explore adding a second augmentation to one or both
of the window copies. The final approach (that was submitted to the challenge)
performs the following augmentations: the first copy of the window is high-pass
filtered at $250$ Hz ($f_1$), followed by rewind with a $0.5$ probability
($f_2$), followed by inverting with a $0.5$ probability ($f_3$), while the
second copy is polluted with uniform noise with a range of $0.02$ ($g_1$) followed by
upsampling with a $0.5$ probability ($g_2$).

We present the results for this setup on the first row of Table
\ref{tab:results_final}, while also reporting the following additional metrics:
F1-score, accuracy, and weighted accuracy. It should be noted that accuracy is
weighed based on the challenge guidelines (for murmur: $\text{``present''}=5$,
$\text{``unknown''}=3$, and $\text{``absent''}=1$, for outcome:
$\text{``abnormal''}=5$ and $\text{``normal''}=1$). In the following two lines
we present similar results where we omit one augmentation from each channel
(rewind and inversion are removed together). Finally, the last line reports
baseline results where no SSL is used at all, and the network is trained from
top to bottom (no frozen weights) and individually for the murmur and outcome
tasks, solely on the 2022 dataset.

The challenge is ranked based on murmur weighted accuracy and clinical outcome
cost. Our models are able to surpass the fully supervised baseline and achieve
a weighted accuracy of $0.668$ for murmur classification and a cost of $10,541$ for
clinical outcome prediction. On the challenge's hidden test set, we scored $0.737$
(Table \ref{tab:murmur-scores})
for murmur weighted accuracy and $11,946$ (Table \ref{tab:outcome-scores})
for outcome cost, achieving the 13th and 7th ranks respectively. Our results on the hidden test set
suggest that our SSL-based model generalizes well to unseen data.


\begin{table}[tbp]
    \centering
    \begin{tabular}{r|r|r|r}
        Training        & Validation & Test & Ranking \\\hline
         0.786 &       0.671 & 0.737 &  13/40 \\\hline
    \end{tabular}
    \caption{Weighted accuracy metric scores and ranking for our final selected entry (team Listen2YourHeart) for the murmur detection task.}
    \label{tab:murmur-scores}
\end{table}

\begin{table}[tbp]
    \centering
    \begin{tabular}{r|r|r|r}
        Training        & Validation & Test  & Ranking \\\hline
        9720 &      10821 & 11946 &  7/39 \\\hline
    \end{tabular}
    \caption{Cost metric scores and ranking, for our final selected entry (team Listen2YourHeart) for the clinical outcome identification task.}
    \label{tab:outcome-scores}
\end{table}

\section{Conclusions}

This work approaches the problem of PCG classification through SSL and
evaluates its performance on the PhysioNet 2022 challenge. Results indicate
that a backbone network trained via contrastive SSL improves performance
compared to end-to-end supervised network training. By exploring and combining
several augmentation techniques we are able to gain some intuition into which
data transformations assist model training in this particular signal domain.
Following this research, we aim to explore additional transformations and to
introduce attention mechanisms into the model.


\bibliographystyle{cinc}
\bibliography{references}


\balance

\end{document}